\documentclass{article}

\usepackage{microtype}
\usepackage{graphicx}
\usepackage{subfigure}
\usepackage{tikz}
\usepackage{lipsum}
\usepackage{adjustbox}
\usepackage{booktabs} % for professional tables

\usepackage{hyperref}

\usepackage[accepted]{icml2025}

\usepackage{amsmath}
\usepackage{amssymb}
\usepackage{mathtools}
\usepackage{amsthm}
\usepackage{tikz-cd}

\usepackage[capitalize,noabbrev]{cleveref}
\usetikzlibrary{shapes.geometric, arrows}

\theoremstyle{plain}

\theoremstyle{definition}

\theoremstyle{remark}

\tikzstyle{trapezoid_upright} = [trapezium, trapezium angle=60, minimum height=1.2cm, draw=black, fill=white]
\tikzstyle{trapezoid_inverted} = [trapezium, trapezium angle=-60, minimum height=1.2cm, draw=black, fill=white]
\tikzstyle{arrow} = [thick,->,>=stealth]

\usepackage[textsize=tiny]{todonotes}

\icmltitlerunning{Submission and Formatting Instructions for ICML 2025}

\begin{document}

\twocolumn[
\icmltitle{Fokker-Planck to Callan-Symanzik: evolution of weight matrices under training}

\begin{icmlauthorlist}
\icmlauthor{Wei Bu}{yyy,xxx}
\icmlauthor{Uri Kol}{zzz}
\icmlauthor{Ziming Liu}{hhh}
\end{icmlauthorlist}

\icmlaffiliation{xxx}{Northeastern University}
\icmlaffiliation{yyy}{Harvard University}
\icmlaffiliation{zzz}{Center of Mathematical Science and Applications, Harvard University}
\icmlaffiliation{hhh}{Massachusetts Institute of Technology}

% \icmlcorrespondingauthor{Wei Bu}{wei\_bu@fas.harvard.edu}


\vskip 0.3in
]



\printAffiliationsAndNotice{}  % leave blank if no need to mention equal contribution

\begin{abstract}
The dynamical evolution of a neural network during training has been an incredibly fascinating subject of study. First principal derivation of generic evolution of variables in statistical physics systems has proved useful when used to describe training dynamics conceptually, which in practice means numerically solving equations such as Fokker-Planck equation. Simulating entire networks inevitably runs into the curse of dimensionality. In this paper, we utilize Fokker-Planck to simulate the probability density evolution of individual weight matrices in the bottleneck layers of a simple 2-bottleneck-layered auto-encoder and compare the theoretical evolutions against the empirical ones by examining the output data distributions. We also derive physically relevant partial differential equations such as Callan-Symanzik and Kardar-Parisi-Zhang equations from the dynamical equation we have. 

%The repulsive distribution of eigenvalues/singular values on the potential is observed, a subset of them escape the original unique Gaussian minima during training. 
\end{abstract}

\section{Introduction}

Understanding the dynamics of neural networks during training has become a focal point in machine learning, offering a rich interplay between optimization, stochastic analysis, and geometry. Recent advances have drawn from statistical physics to conceptualize training dynamics through stochastic differential equations, such as the Langevin equation, and their corresponding evolution equations for probability densities, such as the Fokker-Planck equation. These frameworks provide powerful tools for analyzing stochastic optimization methods like stochastic gradient descent (SGD) and its variants \cite{welling2011bayesian,ye2017langevin,kunin2021limiting,difonzo2022stochastic,bras2022langevin,adhikari2023machine}.

These stochastic differential equations governing Brownian processes have been extensively studied in the context of generative models and score-based reverse diffusion processes. In these cases, the evolution of input data, such as probability densities over high-dimensional input data spaces, has been the primary subject of investigation~\cite{song2021maximum,song2021score,batzolis2021conditional,li2024accelerating}. Techniques like dynamic optimal transport~\cite{lonardi2020designing,haasler2021scalable,tong2023minibatch,kornilov2024optimal} have demonstrated the power of these methods to capture the dynamics of data distributions. However, the evolution of neural network weight matrices themselves during training—a complementary perspective—remains less explored. 

It has been argued that one could design generative models from any partial differential equations (PDE) describing physical processes \cite{liu2023genphysphysicalprocessesgenerative}, and solving these PDEs numerically up to extremely high precision is equivalent to actually training the network to learn the underlying physical processes. In addition to the usual diffusion models \cite{ho2020denoisingdiffusionprobabilisticmodels,song2021maximum,song2021score}, stochastic equations such as Fokker-Planck equation can also be used in physics informed neural networks (PINN) \cite{raissi2019physics,karniadakis2021physics,cuomo2022scientificmachinelearningphysicsinformed,matthews2024pinn}, where the entire network simulates the underlying physical PDE. As one scales up the network, the numerical method requires increasingly more computational power. In the meantime, the neural network encounters the curse of dimensionality as it is an equivalent problem to train these networks simulating solutions to the physical PDEs. This way of designing network training dynamics regards the entire network as a single complicated probability density distribution/wave-packet, and evolve the black box through the physical PDE through discrete training time. In order to disentangle the highly coupled dynamical system, we instead consider a single layer in simple architectures, and try to understand the evolution of the probability density controlling their individual weight matrices through training.

\paragraph{Contributions:}
In this paper, we make an attempt to describe the training dynamics of individual weight matrices. Specifically, we focus on the bottleneck layers of a two-bottleneck-layered auto-encoder, a model that has become central in unsupervised learning and generative modeling~\cite{hinton2006reducing, kingma2014auto, doersch2016tutorial}. Autoencoders, by design, are effective at learning low-dimensional representations of high-dimensional data, making them ideal for exploring the behavior of weight distributions in constrained architectures through explicit visualizations. Variational autoencoders (VAEs) further extend this by modeling the data distribution in a probabilistic framework, making them a natural choice for capturing the stochastic dynamics of weight evolution during training~\cite{kingma2014auto, rezende2014stochastic}. 

Using the Fokker-Planck equation, we simulate the probability density evolution of individual weight matrices, by feeding the same data input as the actual network into the theoretically computed weight matrices, we compare the output data distributions against the actual network output data distribution.

Our work also draws connections to well-established physical frameworks by deriving physically relevant partial differential equations, such as the Callan-Symanzik equation and the Kardar-Parisi-Zhang (KPZ) equation~\cite{KPZ1986}, from the underlying dynamics of weight evolution. These equations, traditionally studied in quantum field theory and surface growth phenomena, respectively, provide a new perspective to interpret neural network training.

% auto-encoder \cite{hinton2006reducing,vincent2008extracting,kingma2014auto,rezende2014stochastic,coates2011unsupervised,makhzani2016adversarial}

% fokker-planck papers \cite{wang2024tensor,rao2023applications}

% Langevin equation papers \cite{welling2011bayesian,zhang2017langevin,kunin2021limiting,difonzo2022stochastic,adhikari2023machine,bras2022langevin}

% Optimal transport and flow matching \cite{kornilov2024optimal,tong2023conditional,tong2023minibatch,lipman2023equivariant,liu2022rectified,haasler2021scalable,lonardi2020designing}

\subsection{Related work}
Studying training dynamics and evolution of the network with statistical physics ideas and stochastic equations is certainly not a novel idea, a number of approaches have been considered before. We list a few of these directions in this section, we apologize in advance for missing out on the references.

\paragraph{Dynamical mean field theory}
The use of dynamical mean field theory (DMFT) \cite{Mei_2018,chizat2018globalconvergencegradientdescent,Sirignano_2018,Rotskoff_2022} in the context of neural networks has evolved from being applicable to the infinite width limit (in the neural tangent kernel (NTK) regime) \cite{arora2019exactcomputationinfinitelywide,jacot2020neuraltangentkernelconvergence,Lee_2020,yang2021tensorprogramsiibarchitectural} to being able to pick up non-trivial features of the data distribution \cite{mei2019mean,Geiger_2020,yang2022featurelearninginfinitewidthneural,bordelon2023dynamicsfinitewidthkernel} and more recently to the finite width limit \cite{Bordelon2023selfconsistent,Zou_2024}. All these models utilize the mean field theory approach to model the collective behaviors of the entire network (usually 2 layered) averaged over different models drawn from a certain joint distribution across all the layer variables. Our work instead focuses on the evolution of a single layer inside a network.

\paragraph{Renormalization in neural network-field theory correspondence}
The use of quantum field theory techniques from physics has spawned the neural network-field theory correspondence, where meaningful observables in a neural network, namely the averaged moments of several neurons are analogous to correlation functions in the context of quantum field theories \cite{Halverson_2021}. Other subsequent papers have migrated other ideas such as renormalization group flow in field theories to neural networks given the analogy \cite{erbin2022renormalizationneuralnetworkquantumfield,Howard:2024kfd, Cotler:2022fze,Cotler:2023lem}. The scale involved in the renormalization flow of these models are neuron distributions, whereas in the Callan-Symanzik equation we derive, the role of the scale is played by training time.

%%%%%%%%%%%%%%%%%%%%%%%%%%%
\section{Analytic side}
\subsection{Technical background}
To describe the dynamical evolution of a single weight matrix, we first need to identify the meaningful physical quantity that define the weight matrix. For instance, in a teacher-student setting, this can be the dot product between the teacher (known ideal optimal) weight matrix and student weight matrix. Another natural choice is inherited from random matrix theory \cite{Potters_Bouchaud_2020}, where one can see all the entries of weight matrix as drawn from a certain 1d probability distribution $P$, which can be described by a effective potential function:
\begin{equation}
    V = -\log{P}
\end{equation}
For example, simple Gaussian distribution can be described by a $V=x^2$ potential with a unique minima. 

Alternatively, one could also choose to view matrices of dimension $m\times n$ ($m$ rows and $n$ columns) as $m$ sets of points on $\mathbb{R}^n$. Since matrices symbolize transformations of vectors in $\mathbb{R}^n$, viewing matrices this way allows one to see the operation of the matrix on input data vector explicitly. For example, for a $200\times 2$ matrix, instead of seeing the 1d distribution for all matrix entries as the physical quantity, one could study the 2d distribution the $200$ samples were drawn from. Similar to the 1d perspective, this allows one to write down $P(x_1,x_2)$ from which $200$ independent draws can be made to give the rows of the weight matrix. We will show that this indeed defines a high dimensional probability density with total probability integrated to $1$, which remains constant during training. The effective potential can be defined in a similar fashion:
\begin{equation}\label{Potential-probability}
    V(x_1,x_2) = -\log(P(x_1,x_2))
\end{equation}
For example a 2d Gaussian distribution would contribute to
\begin{equation}
    P(\Vec{x}) = \frac{1}{2\pi|\Sigma|^{1/2}}\,\text{exp}\left(-\frac{1}{2}(\Vec{x}-\Vec{\mu})^{\text{T}}\Sigma^{-1}(\Vec{x}-\Vec{\mu})) \right) 
\end{equation}
where $\Vec{x}=(x_1,x_2)$, $\Vec{\mu} = (\mu_1,\mu_2)= \frac{1}{m}(\sum_{i=1}^{m} x_{i1},\sum_{i=1}^{m} x_{i2})$ is the mean vector with each component averaged over all the $m=200$ row elements. 
\begin{multline}
    \Sigma = \frac{1}{m}\sum_{k=1}^{m}(\Vec{x}_{k}-\Vec{\mu})(\Vec{x}_k-\Vec{\mu})^{\text{T}}=\\
    \frac{1}{m}\sum_{k=1}^m
    \begin{pmatrix}
        (x_{k1}-\mu_1)^2 & (x_{k1}-\mu_1)(x_{k2}-\mu_2)\\
        (x_{k1}-\mu_1)(x_{k2}-\mu_2) & 
        (x_{k2}-\mu_2)^2
    \end{pmatrix}
\end{multline}
labels the covariance matrix. For a simple zero mean $\Vec{\mu}= (0,0)$, covariance matrix $\Sigma=\frac{1}{2} \begin{pmatrix}
    1 & 0\\
    0 & 1
\end{pmatrix}$, the potential is just 
\begin{equation}
    V(\Vec{x}) = x_1^2+x_2^2
\end{equation}
Freezing training evolution at any point, each weight matrix in a neural network is just a matrix, which can be seen as a manifestation of an effective potential. This high dimensional potential (the same as column dimension of the weight matrix) evolves during training, we would like to write down an equation to track its evolution.

\medskip

\paragraph{Stochastic differential equations}

Stochastic diffusion equation as a basic first principal assumption of evolution of a generic dynamical variable. For a generic weight matrix row $\Vec{w}$, a time-dependent It\^{o} process:
\begin{equation}\label{SGD_equation}
    d \Vec{w} = \Vec{D}(\Vec{w},t)d t + \Vec{\sigma}(\Vec{w},t)\, d \Vec{B}_t
\end{equation}
where $\Vec{D}$ is the deterministic drift vector, $\Vec{\sigma}$ denotes the diffusion coefficients and $d \Vec{B}_t$ is assumed to be a average $0$ high dimensional Wiener process:
\begin{equation}
    \langle \Vec{B}_t\rangle = 0 \,, \quad \langle \Vec{B}_{t_1}(\Vec{w}_1)\Vec{B}_{t_2}(\Vec{w}_2)\rangle = \delta(\Vec{w}_1-\Vec{w}_2)\delta(t_1-t_2)
\end{equation}
where the averages are taken against any probability density distribution, $\delta$ denotes the familiar Kronecker delta function. The source of stochasticity is presumed to come from some unknown processes, it appears as a part of the assumption of the theory to account for currently unforeseeable randomness.

Entries of $\Vec{\sigma}(\Vec{w},t)>0$ are the diffusion coefficients describing the strength of random walk. For the deterministic part of the evolution, since we are trying to model a training process, it is important to update the weights according the direction where loss is descending. For this purpose we set a generic form for $\Vec{D}(\Vec{w},t)$:
\begin{equation}\label{Drift_coefficient}
    \Vec{D}(\Vec{w},t) = H\left(-\frac{\partial L(\Vec{w},t)}{\partial t}\right)\, \Vec{F}(\Vec{w},t)
\end{equation}
with $H\left(-\frac{\partial L(\Vec{w},t)}{\partial t}\right)$ the usual Heaviside step function. Physically it only allows for the deterministic evolution of $\Vec{w}$ when the loss is decreasing\footnote{Although the definition of $\Vec{F}$ involves the loss through the optimizer choice as we shall see, we learn from practice that the use of an optimizer does not guarantee the loss to be decreasing.}. Components of $\Vec{F}(\Vec{w},t)$ are update rules informed by the actual optimizer implemented. In our auto-encoder context, this is chosen to be ADAM\footnote{One is free to choose different optimizers according to the architecture.}.

In actual implementation, the time step is taken to be $1$ epoch, in which case $d t =1$ in equation \eqref{SGD_equation}, all the components of $\Vec{w}$ follow the weight update rule:
\begin{equation}
    w_{t+1} = w_t \underbrace{-\frac{\eta}{\epsilon+\sqrt{\hat{v}_{t+1}}}\,\hat{m}_{t+1}}_{\text{each component of}\Vec{F}(w,t)}
\end{equation}
where $\hat{v}_{t+1}=\frac{\beta_2 v_t+(1-\beta_2)g(w_t)^2}{(1-\beta_2^{t+1})}$ is the renormalized velocity, $\hat{m}_{t+1}=\frac{\beta_1 m_t+(1-\beta_1)g(w_t)}{(1-\beta_1^{t+1})} $ is the renormalized momentum. $g(w)$ here is the gradient of the loss function with $\beta_1$ and $\beta_2$ some parameter between $0$ and $1$. 

Each weight matrix can be thought of as being generated from some effective probability distribution $P(\Vec{w},t)$ evolving as training epoch progresses. One could model the time evolution of $P(\Vec{w},t)$ by examining the expectation of some generic function of $f$ against the probability distribution, which can be denoted 
\begin{equation}
    \left\langle f(\Vec{w},t)\right\rangle = \mathbb{E}_{P}[f(\Vec{w},t)] = \int f(\Vec{w},t)P(\Vec{w},t) d \Vec{w}
\end{equation}
With the stochastic diffusion equation \eqref{SGD_equation}, one can obtain the single time step evolution of $\langle f(\Vec{w},t)\rangle$. Integration by parts gives one the evolution of the probability density inside the one-point expectation value of the test function $f$:
\begin{multline}\label{FP_equation}
    \frac{\partial P(\Vec{w},t)}{\partial t}= \\
 -\nabla\cdot\left(\Vec{D}(\Vec{w},t) P(\Vec{w},t)\right) + \frac{1}{2}\,\Delta\left(\Vec{\sigma}^2 P(\Vec{w},t)\right)
\end{multline}
where components of $\Vec{w}$ is an weight entry in the matrix, $\nabla\cdot$ denotes the divergence with respect to all components of $\Vec{w}$ while $\Delta = \nabla\cdot \nabla \cdot $ is its Laplacian. Here we used $\Vec{\sigma}^2=\Vec{\sigma}^{\text{T}}\Vec{\sigma}$ to label a diffusion coefficient matrix.

$\Vec{\sigma}$ quantifies the variance $\varepsilon$ of the ADAM optimizer together with the learning rate $\eta$:
\begin{equation}\label{diffusion_coefficient}
    \vec{\sigma}^2\sim \text{diag.}(\varepsilon^2\eta^2)
\end{equation}
The two terms on the right hand side compete with each other, the drift term deterministically move the probability density towards the training goal. In the meantime, the diffusion term randomly pulls the density in all directions. 

One could exchange this for the equation for the effective potential $V(\Vec{w},t)$ governing the probability density distribution by substituting in \eqref{Potential-probability}:
\begin{multline}
    \frac{\partial V}{\partial t} = \sum_{i,j}\frac{{\sigma}^2_{ij}}{2}\frac{\partial V}{\partial w_i}\frac{\partial V}{\partial w_j} - \nabla\cdot\left(\frac{\Vec{\sigma}^2}{2}\nabla V \right)\\
    + \nabla V \cdot \Vec{D}+\nabla\cdot \Vec{D}-\frac{1}{2}\Delta \Vec{\sigma}^2
\end{multline}
the more familiar 1d version can be written as:
\begin{multline}
    \frac{\partial V}{\partial t} = \frac{\sigma^2}{2}\,\frac{\partial^2 V}{\partial w^2}-\frac{\sigma^2}{2}\,\left(\frac{\partial V}{\partial w}\right)^2\\
    +\frac{\partial V}{\partial w}\left(\frac{\partial\sigma^2}{\partial w}-D\right)+\left(\frac{\partial D}{\partial w}-\frac{1}{2}\frac{\partial^2\sigma^2}{\partial w^2}\right)
\end{multline}
which is a generalization of the celebrated Kardar-Parisi-Zhang (KPZ) equation \cite{KPZ1986}. The additional term that exists is the linear drift term $\frac{\partial V}{\partial w}\left(\frac{\partial\sigma^2}{\partial w}-D\right)$. In the KPZ equation, the constant term in the equation should be a randomized function with zero average and delta function like variance, whereas here the constant term $\left(\frac{\partial D}{\partial w}-\frac{1}{2}\frac{\partial^2\sigma^2}{\partial w^2}\right)$, which behaves like a combination of drift and diffusion.

\subsubsection{Connection to Physical systems}
If one were to ignore the diffusion term in \eqref{FP_equation} and further rewrite on the drift function $\Vec{D}(\Vec{w},t)$ in a specific form:
\begin{equation}
    \Vec{D}(\Vec{w},t) = \Vec{G}(\Vec{w},t)/t
\end{equation}
where $\Vec{G}(\Vec{w},t) =H\left(-\frac{\partial L}{\partial t}\right)\Vec{F}(\Vec{w},t)\,t$ is some function with a first order $0$ at $t=0$. Then the evolution equation can be rewritten as 
\begin{equation}
    \Vec{G}\cdot\nabla P + t\,\frac{\partial P}{\partial t} + P\,\nabla\cdot \Vec{G}  = 0
\end{equation}
If we were to redefine a few coefficient functions, this can simply be written as 
\begin{equation}\label{CS_equation}
    \Vec{\beta}(\Vec{w},t)\cdot\nabla P + t\,\frac{\partial P}{\partial t} + n(\Vec{w},t) P = 0
\end{equation}
where $\Vec{\beta}(\Vec{w},t)=\Vec{G}(\Vec{w},t)$, $n(\Vec{w},t)=\nabla\cdot \Vec{G}$. \eqref{CS_equation} is a vector form of the Callan-Symanzik equation \cite{Callan1970,Symanzik1970}, which describes how an observable such as a probability density distribution evolves deterministically under a scale change in the system (which in this case is training epoch $t$). The function $\Vec{\beta}(\Vec{w},t)$ is the "coupling" term, which indicates how the distribution changes with "coupling", in this case it is the weight matrix rows $\Vec{w}$. Compared to usual $\Vec{\beta}$ function defined as 
\begin{equation}
    \Vec{\beta} = t\,\frac{\partial \Vec{w}}{\partial t}
\end{equation}
We see this is indeed the case for our $\Vec{\beta}(\Vec{w},t)$ given the definition of the drift $\Vec{D}$:
\begin{equation}
    \Vec{D}(\Vec{w},t) = \frac{\Vec{\beta}(\Vec{w},t)}{t}
\end{equation}
which in the current context of training a neural network, describes how weight matrices are updated.

The second term in the equation can be interpreted in the following manner. For a homogeneous function $P(\Vec{w},t)\sim t^m, m\in\mathbb{Z}$ in $t$, $t\,\frac{\partial}{\partial t}$ measures the homogeneity $m$ of $P(\Vec{w},t)$:
\begin{equation}
    t\,\frac{\partial }{\partial t} t^m = m\,t^m
\end{equation}
The same applies to a generic function of $t$, hence $t\,\frac{\partial }{\partial t} P$ physically describes how the distribution $P$ changes under scaling of training time. Callan-Symanzik equation is traditionally used to describe how a system evolves under a change of scale, termed "renormalisation group flow" in physics, the beta function $\Vec{\beta}(\Vec{w},t)$ controls the behavior of the system as a certain scale $t$ changes. Importantly, in certain physical systems such as Quantum Chromodynamics (QCD), the vanishing of the $\Vec{\beta}$ function during the evolution indicates a point where the physical system becomes scale invariant, making it completely manageable analytically. 

In equation \eqref{CS_equation}, this corresponds to when $t\,\frac{\partial P}{\partial t}=0$. The analogous statement in our case will be when training time parameter update no longer affects the probability density distribution of the weight matrix, which reaches a fixed point in training time. Note that either 
\begin{equation}\label{eq:conditions}
    \left.t\,\frac{\partial P}{\partial t} =0 \right\vert_{t=T}\text{ or } \left.\frac{\partial P}{\partial t}= 0\right\vert_{t=T}
\end{equation}
would eliminate the second term from equation \eqref{CS_equation}, where we denoted the training epoch satisfying such conditions $T$. The former indicates $P$ being invariant under rescaling of $t$\footnote{In other words, training with multiples of a certain time scale $T$ would not effectively affect the distribution controlling the weight matrix.}, the latter simply suggests a stationary point of the probability density distribution during training. In this case the flow equation is completely controlled by the $\Vec{\beta}$-function or $\Vec{G}(\Vec{w},t)$. $P$ is given by 
\begin{equation}\label{Stationary_equation}
    \nabla\cdot\left(P\Vec{G}\right) = 0
\end{equation}
In order to solve for $\left. P(\Vec{w},t)\right\vert_{t=T}$ at terminal time $T$, we can expand the left hand side and write down some parametrized high dimensional curve $\Vec{w}(s)$ with $s$ the coordinate along the curve such that $\left.\Vec{G}(\Vec{w},t)\right\vert_{t=T} = \frac{d \Vec{w}}{d s}$, this allows one to write the equation as
\begin{equation}\label{Terminal_solution_highd}
\begin{aligned}
    &\frac{d P(\Vec{w}(s))}{d s} = -P(\Vec{w}(s))\left(\nabla\cdot \Vec{G}(\Vec{w}(s))\right) \implies \\
    & P(\Vec{w}(s)) = P_0\,\text{exp}\left(-\int^s_0\left(\nabla\cdot \Vec{G}\right)(\Vec{w}(s')) d s'\right)
\end{aligned}
\end{equation}
where $P_0$ is determined by the initial distribution. \eqref{Stationary_equation} also implies the conservation of some probability moderated drift charge: 
\begin{equation}
    \oint_{\partial A} (P\Vec{G})\cdot\Vec{n} \,d S =0
\end{equation}
where $\partial A$ is the boundary of some closed hypersurface $A$, $\Vec{n}$ its normal vector and $dS$ the area element. With this, one could integrate both sides of \eqref{CS_equation} and obtain
\begin{equation}
    \frac{d}{d t}\int_{\mathbb{R}^n} P(\Vec{w},t)  =  -\oint_{\partial \mathbb{R}^n} (P\Vec{G})\cdot\Vec{n} \,d S = 0 
\end{equation}
where we used Gauss law on the right hand side with $\partial \mathbb{R}^n$ a infinite radius hyper-sphere $S^{n-1}$. This indicates the fact that the total probability $\int_{\mathbb{R}^n} P(\Vec{w},t)$ is constant over training time, one could choose this to be $1$, hence legitimizing the definition of $P(\Vec{w},t)$ as a probability density.

In the 1d case, or when all components of the drift vector are decoupled, one can write down the following simple solution: 
\begin{equation}\label{Terminal_solution_1d}
    \left.P(w,t)\right\vert_{t= T} = \left.\frac{\text{const.}}{G(w,t)}\right\vert_{t=T}
\end{equation}
where $T$ is the terminal training time where either condition in \eqref{eq:conditions} is met.
It would be both theoretically and pragmatically interesting to further study this analog and study the implications of equation \eqref{Terminal_solution_highd}. We make two remarks here. Equation \eqref{Terminal_solution_highd} suggests a computable probability density distribution controlling individual weight matrices under either condition \eqref{eq:conditions}. It is determined by how one chooses to update the parameters in the network \eqref{Drift_coefficient}. On the other hand, it is unclear how to distinguish whether the probability density computed this way is in fact the end of training (where the probability density no longer updates) or just a time scale invariant point. Nonetheless it is describing a quite special phase of training where everything is determined by the drift. This interesting solution \eqref{Terminal_solution_highd} awaits further elucidation.

\paragraph{Practical implications:}
We envision there to be practical implications from this interesting fact. Although the computation of terminal time $T$ requires information from training, one could pre-train the model for a short time, then use the information to extrapolate the value of terminal $T$. This value of $T$ can be used to estimate the terminal training epoch at which the model no longer have meaningful updates (ones that change the distributions behind individual layers). Doing this for each layer in the model could give estimation of terminal training time for complex architectures.

%%%%%%%%%%%%%%%%%%%%%%%%%%%%

%%%%%%%%%%%%%%%%%%%%%%%%%
%%%%%%%%%%%%%%%%%%%%%%%%%

%%%%%%%%%%%%%%%%%%%%%%%%%
%%%%%%%%%%%%%%%%%%%%%%%%
\section{Empirical results}
\subsection{Set up}
We work in a simple auto-encoder set up \ref{fig:architecture} for MNIST (with $0-5$), with the following architecture:
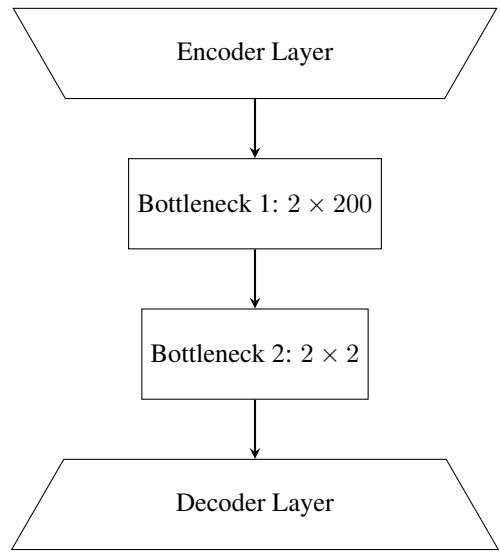
\begin{figure}
    \centering
    \begin{tikzpicture}[scale = 0.6, node distance=2cm]
    % Encoder layers (inverted trapezoids)
    \node (enc1) [trapezoid_inverted, minimum width=4cm] {Encoder Layer};
    % Bottleneck (rectangle)
    \node (bottleneck1) [rectangle, minimum width=2cm, minimum height=1.2cm, draw=black, fill=white, below of=enc1] {Bottleneck 1: $2\times 200$};

    \node (bottleneck2) [rectangle, minimum width=2cm, minimum height=1.2cm, draw=black, fill=white, below of=bottleneck1] {Bottleneck 2: $2\times 2$};
    % Decoder layers (upright trapezoids)
    \node (dec1) [trapezoid_upright, below of=bottleneck2, minimum width=3cm] {Decoder Layer};
    % Arrows
    \draw [arrow] (enc1) -- (bottleneck1);
    \draw [arrow] (bottleneck1) -- (bottleneck2);
    \draw [arrow] (bottleneck2) -- (dec1);
    \end{tikzpicture}
    \caption{Architecture set up}
    \label{fig:architecture}
\end{figure}

where importantly, the bottleneck region has two linear layers of latent dimension 2 with no activation functions. This allows us to visualize the distribution of the potential in the 2 layers in a straightforward manner without further dimensional reduction techniques such as UMAP. 

\paragraph{Main Assumption:}We work with a simple auto-encoder architecture with two 2d bottleneck layers with no activation functions in between them. So in latent space, data space is manifestly 2 dimensional, the two layers are linear within the bottleneck region with two feed-forward weight matrices. This provides the ideal framework to test the hypothesis of whether effective probability density controlling the weight matrices evolves under the full Fokker-Planck type equation \eqref{FP_equation}.

We are using the notion of a 2d dimensional probability density distribution over the two dimensions of the bottleneck weight matrices. More specifically, the PDE we are implementing in practice is 
\begin{equation}
    \frac{\partial P(\Vec{x},t)}{\partial t} = 
    -\nabla\cdot\left(\Vec{D}(\Vec{x},t)P(\Vec{x},t) \right) +\frac{1}{2}\,\Delta\left(\Vec{\sigma}^2P(\Vec{x},t) \right)
\end{equation}
where $\Vec{x}=(x_1,x_2)$ represents the two column dimensions of the weight matrices in the bottleneck layers and $\nabla$ is the gradient with respect to the two components of $\Vec{x}$. 

\begin{figure}[ht]
    \centering

    \adjustbox{scale =0.83, center}{
    \begin{tikzcd}
     &  \text{Input data} \arrow[dl] \arrow[dr]\\
     \text{Bottleneck layers} \arrow[d] & & \text{Theory evolved matrices} \arrow[d]
     \\
     \text{Empirical output data} & & \text{Theory output data}
    \end{tikzcd}
    }
    \caption{Comparison between empirical and theoretical output data distribution}
    \label{fig:method}
\end{figure}
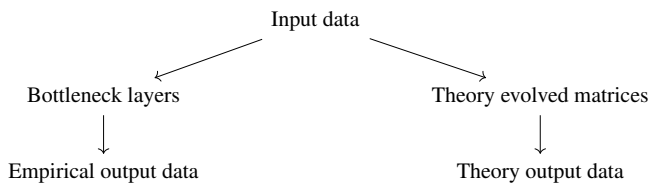

Notice that bottleneck layers have $200$ and $2$ rows respectively, if we were to directly estimate the potential governing their weight matrices, it simply is not quite possible given the lack of sample points. In order to visualize and compare the empirical and theoretical prediction about the weight matrices, in practice, we instead examine the data distribution in latent space after passing through individual layers as pictured in figure \ref{fig:method}. Effectively, we pass input data through both the actual network and the an ensemble of weight matrices given by the potential computed from the theoretical computation and compare the output data distributions. 

%%%%%%%%%%%%%%%%%%%%%%%%%%%%%
\subsection{Interpretation of diagrams}
\begin{figure}[ht]
    \centering
    \includegraphics[width=0.95\linewidth]{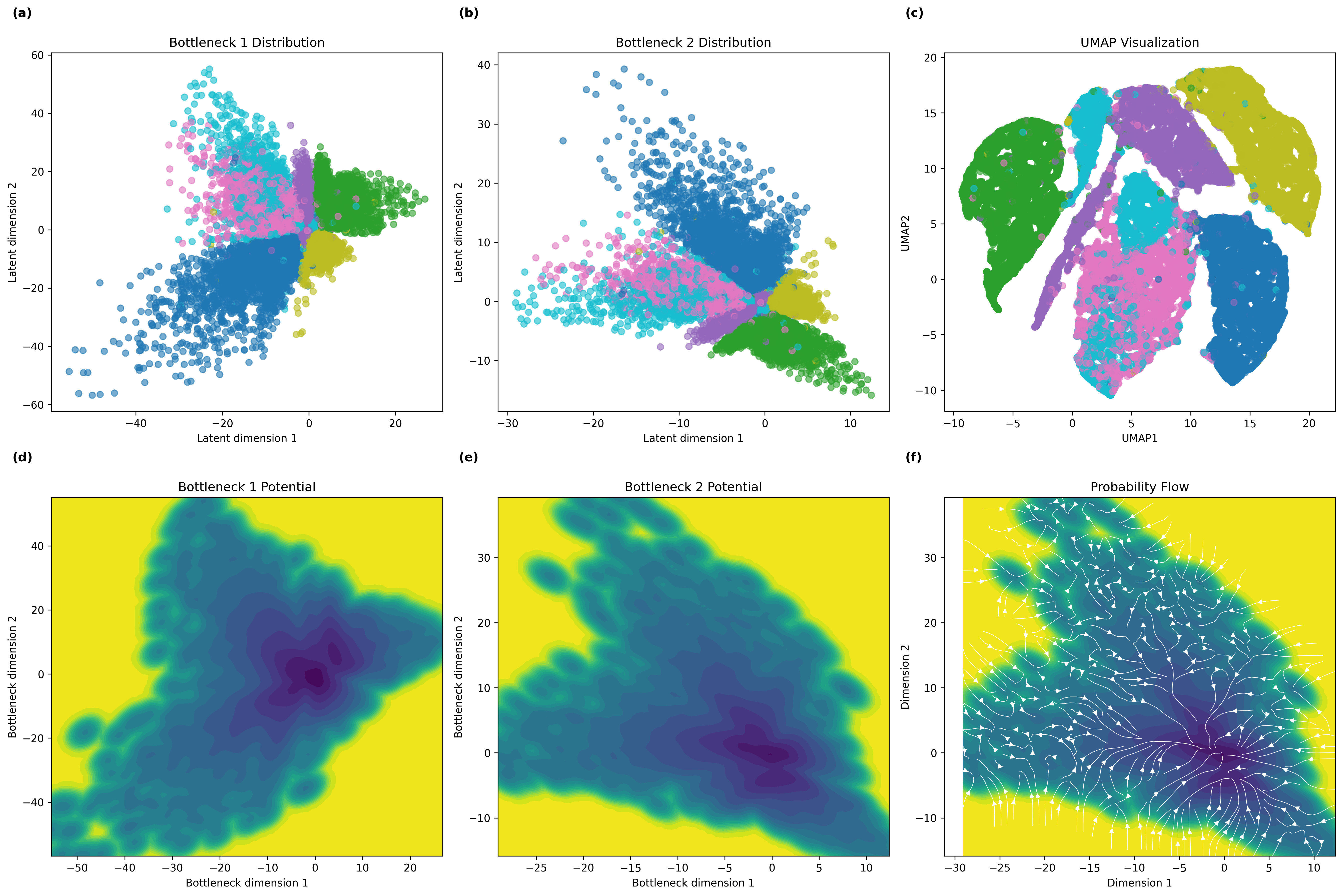}
    \caption{Example output data distributions of bottleneck weight matrices.}
    \label{fig:example_combined}
\end{figure}

In the diagram below figure \ref{fig:example_combined}, $(a)$ and $(b)$ depict how data is distributed after passing through the two bottleneck layers. Each color represents a distinct handwritten digits, the same digits have been drawn with the same color. Correspondingly, $(d)$ and $(e)$ are the kernel density estimated (KDE) 2d potential distributions of data after passing through two bottleneck layers estimated from the scatter diagrams above, where the two axis are taken to be the minimum to the maximum of all the data values. 

Figure $(f)$ shows the direction of gradient on the 2d latent data distribution. The directions of the white arrows point along the potential-descending direction in latent dimensions. This is usually used to define a score function in modern generative models \cite{song2021maximum,song2021score,batzolis2021conditional,li2024accelerating}. 

Finally subfigure $(c)$ in figure \ref{fig:example_combined} demonstrates the UMAP visualization of the actual data manifold after the decoder, which confirms the appropriate clustering of closely related hand-written digits, it serves the purpose of a sanity check for the first two diagrams in the bottleneck layers.

\medskip

In order test our conjecture that the evolution of probability density/effective potential that controls columns of individual weight matrices should follow the equation \eqref{FP_equation}, we compare the output data distributions of the two bottleneck layers against the output data distribution after passing the same input data through the weight matrices controlled by numerically evolved weight space distributions. 

The computation of the weight space distributions follows the following rules. We first use the empirical 2d potential $V_{t_1}$ of the previous epoch $t_1$ to compute the initial condition for the 2d probability density $P_{t_1} =\mathrm{e}^{-V_{t_1}}$. Then we divide the next epoch from $t_1$ to $t_1+1$ into 100 numerical steps. We compute the drift vector $\Vec{D}(\Vec{w},t)$ according to equation \eqref{Drift_coefficient} and the diffusion coefficient $\Vec{\sigma}(\Vec{w},t)$ according to \eqref{diffusion_coefficient}. Then this informs us how to update $P$ recursively according to equation \eqref{FP_equation}, reiterating this 100 times, one covers the entire epoch evolution of the training process. Using this updated weight space probability density $P$, we feed the same input data into weight matrices drawn from it and simply plot the output data distribution. In the diagram below, $(a)$, $(c)$ are the empirical data distributions of the two bottleneck layers (copied from $(d)$ and $(e)$ of figure \ref{fig:example_combined}). $(b)$, $(d)$ are the corresponding numerically evolved data distributions obtained in the way we just described. We compare them by computing the mean-square-error and Pearson correlation defined as 
\begin{equation}
\begin{aligned}
    &\text{MSE} = \frac{1}{n}\,\sum_{i=1}^n\left(y_i-x_i \right)^2\\
    &\text{Correlation} = \frac{\sum(x_i-\bar x)(y_i-\bar y)}{\sqrt{\sum(x_i-\bar x)^2\,\sum(y_i-\bar y)}}
\end{aligned}
\end{equation}
where $x_i$ and $y_i$ are the samples from the empirical and theoretical distributions, $\bar x$, $\bar y$ are their corresponding averages.

\begin{figure}[ht]
    \centering
    \includegraphics[width=0.8\linewidth]{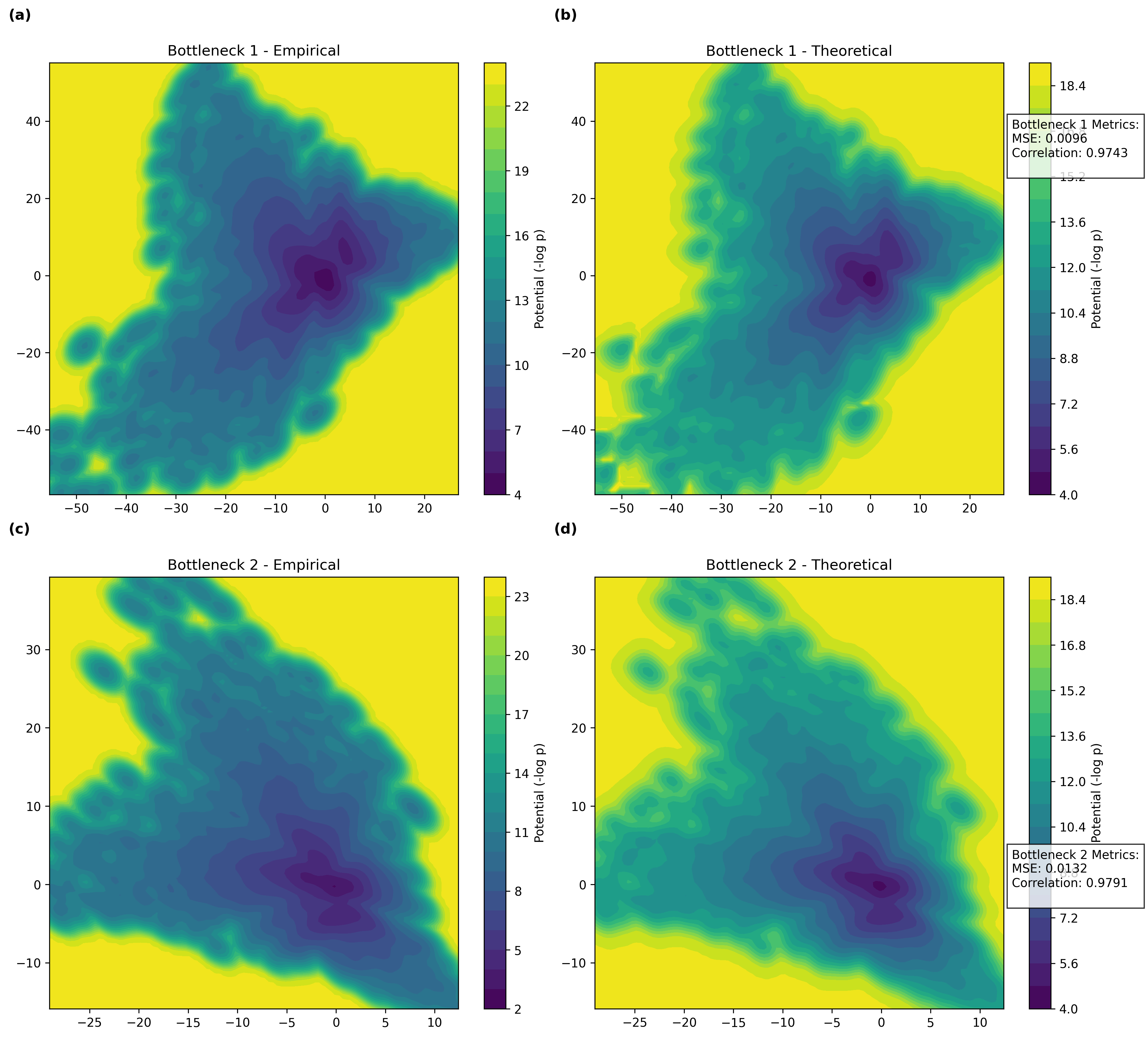}
    \caption{Example of theoretical and empirical comparison of output data distributions during training.}
    \label{fig:FP_compare_example}
\end{figure}

%%%%%%%%%%%%%%%%%%%%%%%%%
%%%%%%%%%%%%%%%%%%%%%%%%%
\subsection{Results}
At initialization (epoch 0), we begin with 2d Gaussian distributions for the data input and in the two bottleneck layers weight space, the same initial condition is given to the Fokker-Planck simulator. 

\begin{figure}[ht]
    \centering
    \includegraphics[width=0.93\linewidth]{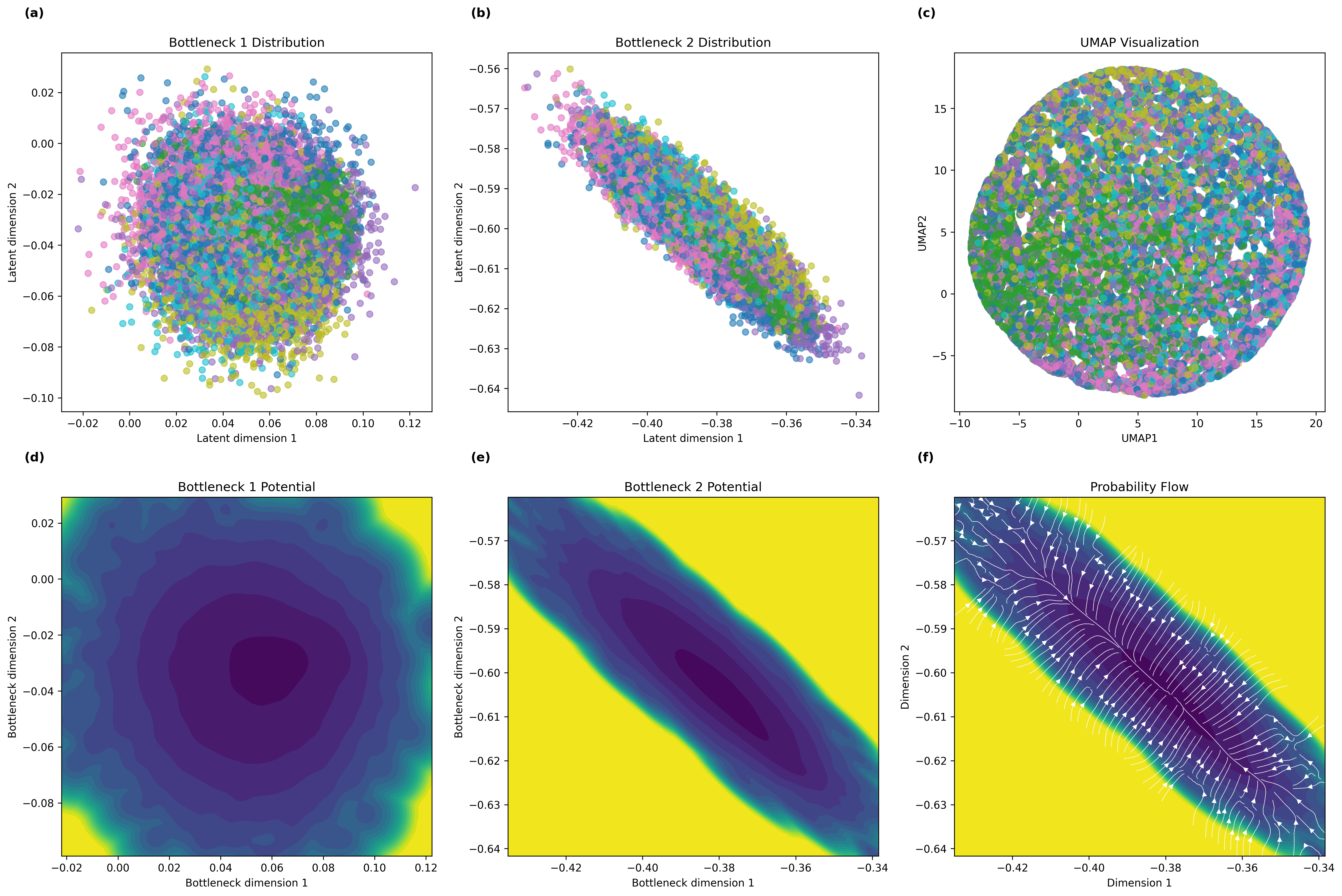}
    \caption{Effective output data distribution plots at initialization}
    \label{fig:combined_0}
\end{figure}
We see that an even mix of all the MNIST data from $0-5$. The estimated output data distribution plots are even in the both directions indicating a Gaussian initialization in figure \ref{fig:combined_0}. 
%The reason $(e)$ is a slightly stretched circle is the variance were set to be slightly different in the two directions when generating weights in bottleneck layer 2.

\begin{figure}[ht]
    \centering
    \includegraphics[width=0.8\linewidth]{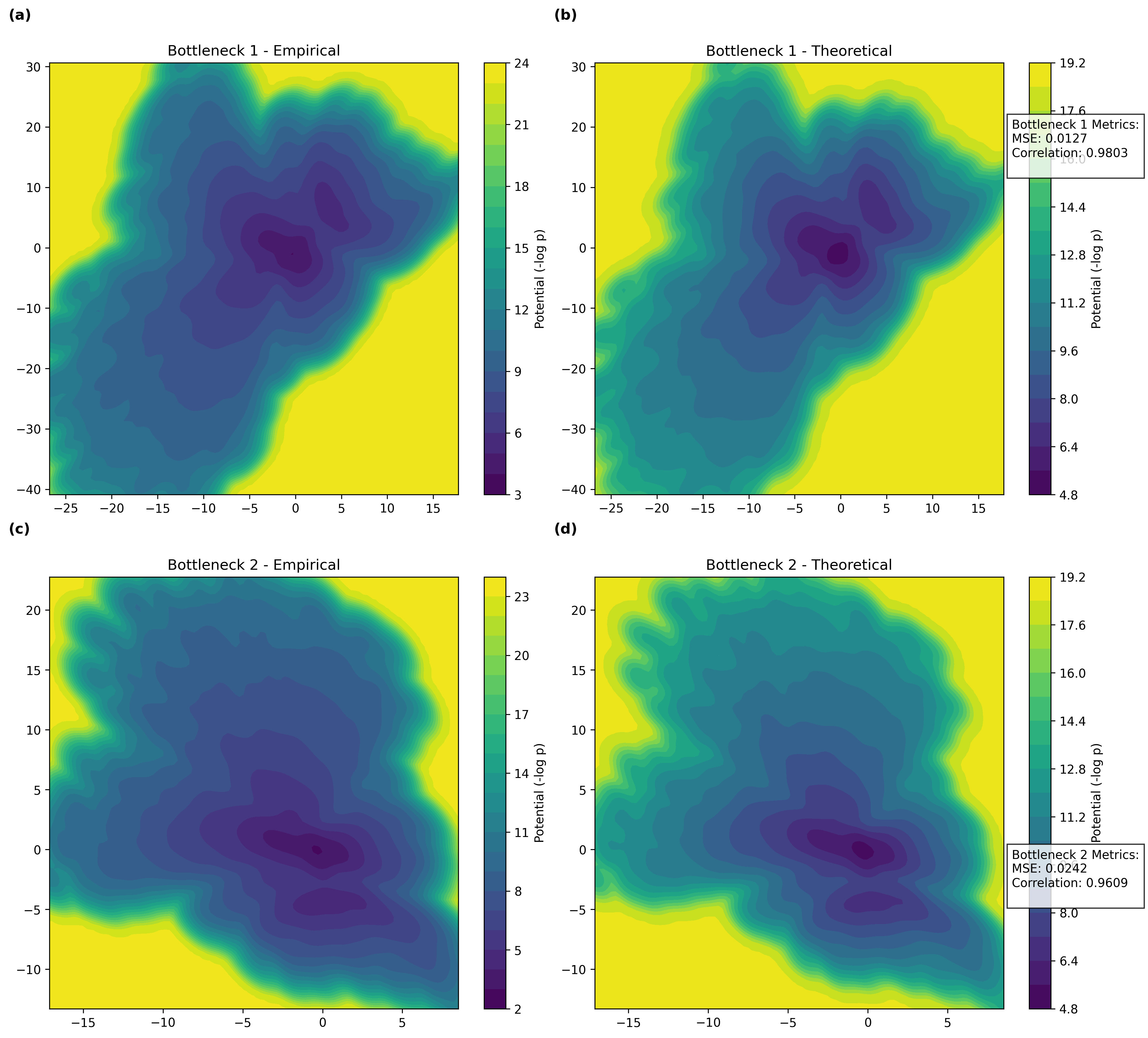}
    \caption{Theoretical vs empirical comparison for output data distributions at epoch 5.}
    \label{fig:FP_compare_5}
\end{figure}

As training epoch progresses, we have the following two diagrams at epoch 5 figure \ref{fig:combined_5} and \ref{fig:FP_compare_5}:
\begin{figure}[ht]
    \centering
    \includegraphics[width=0.95\linewidth]{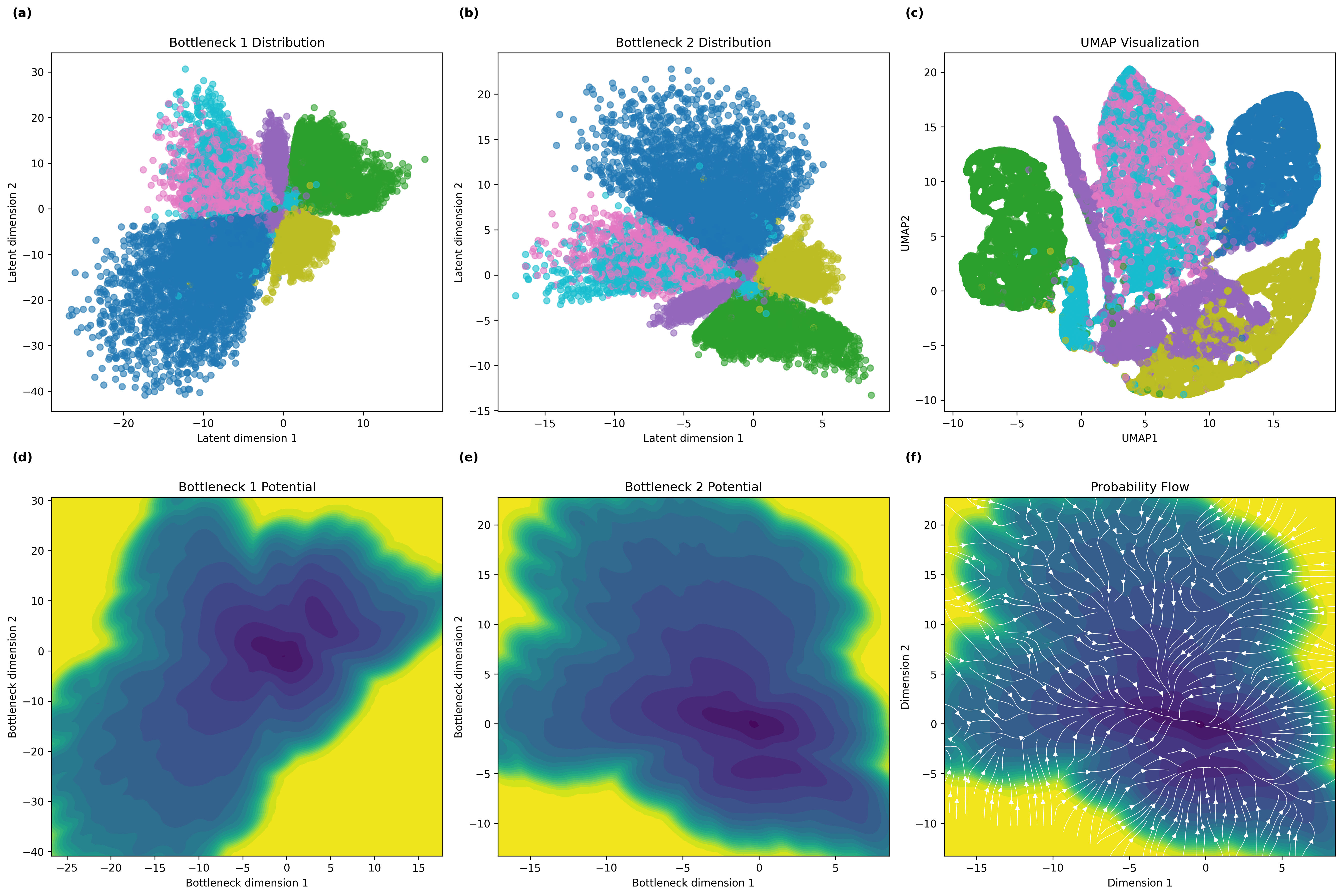}
    \caption{Output data distribution plots at epoch 5.}
    \label{fig:combined_5}
\end{figure}
where we see the expected morphing of the output data distributions, which develops multiple distinct clusters. This deformation of the data manifold latent space distribution is fed-forward through the two layers. The clustering indicates the network is indeed learning the semantic differences and started to classify the hand written digits.

Using the method mentioned before, we simultaneously evolve the weight matrices distributions from Gaussian at epoch 0, this allows us to feed the same data input and compute the output data distribution plotted here at epoch 5 in figure \ref{fig:combined_5}. We see rather low MSE $\sim 0.012$ and rather high Pearson correlation coefficients $\sim 0.98$ in figure $(b)$ of \ref{fig:FP_compare_5}. This shows good approximation of the training evolution using the equation \eqref{FP_equation}. 

\begin{figure}[ht]
    \centering
    \includegraphics[width=0.95\linewidth]{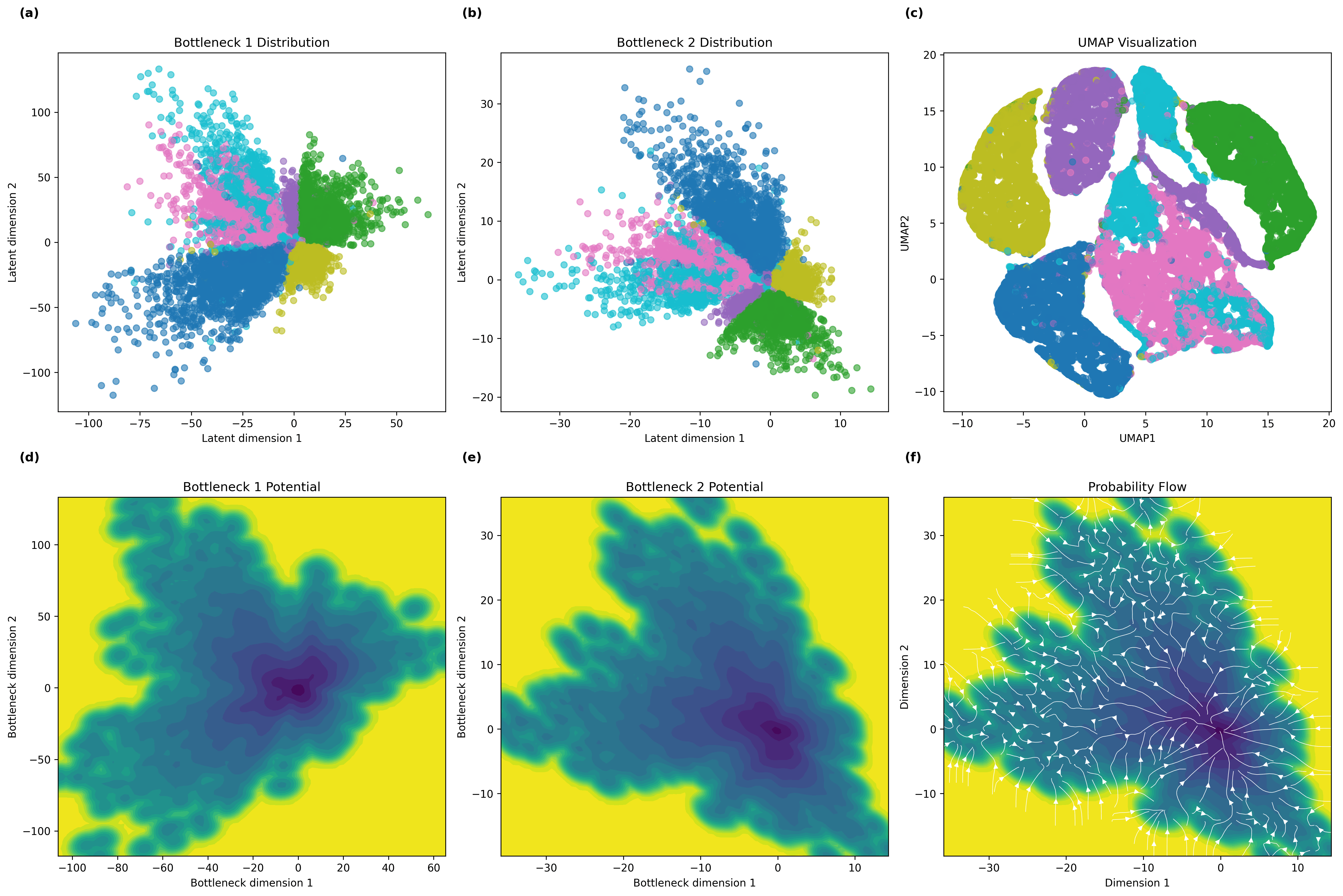}
    \caption{Output data distribution plots at epoch 80.}
    \label{fig:combined_80}
\end{figure}

\begin{figure}[ht]
    \centering
    \includegraphics[width=0.8\linewidth]{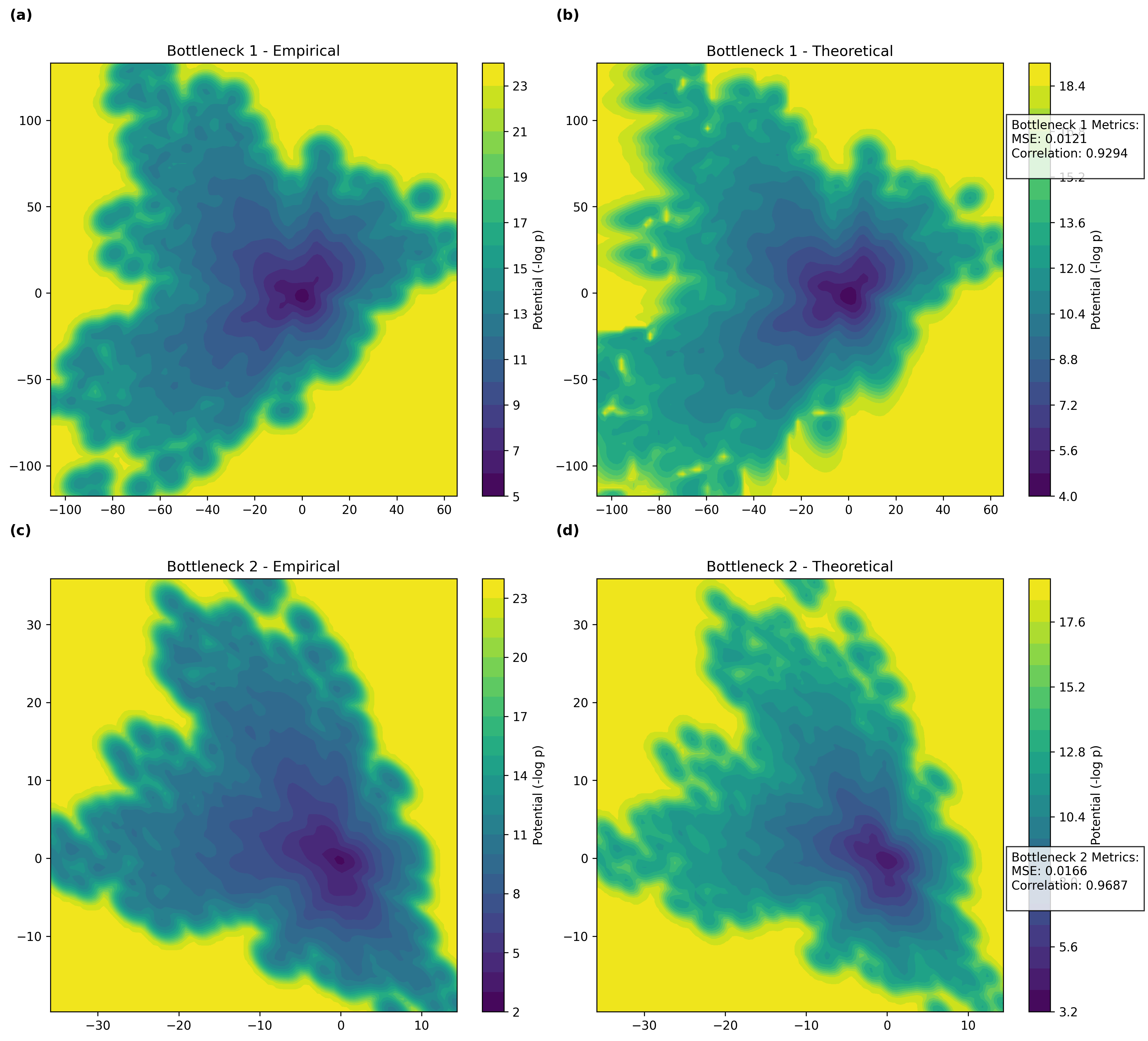}
    \caption{Theoretical vs empirical comparison for output data distributions at epoch 80.}
    \label{fig:FP_compare_80}
\end{figure}

At a further epoch (epoch 80), the loss stabilizes to $\sim 0.03$. We show the same two diagrams in figure \ref{fig:combined_80} and \ref{fig:FP_compare_80}.

This is limited by the architecture we have, with simple 2d latent space and no activation function in the bottleneck region. It is a designed setting for our purpose of easily visualizing and testing the conjecture about evolution of weight matrices under training using the output data distributions. It suffices to see the clustering of data manifold in the layers. If one were to increase the latent space dimension, add activation functions, increase number of layers in the bottleneck region or employ more efficient generative model techniques such as variational auto-encoders, a better loss can be achieved. These more practically relevant architectures require further developments of the coupled dynamics during training without invoking the curse of dimensionality.

The fact that the Pearson correlation coefficients are not exactly $1$ tells us either the theoretical model we are using is only approximating the evolution of the weight matrix, or it is interfered by the following several possible causes:
\begin{itemize}
    \item The diffusion term adding noise to the theoretical computation.
    \item Not enough rows are drawn from the distribution to exactly capture the weight space distribution. 
    \item Random seeds give different initializations, which affects the theoretical computations.
\end{itemize}
%%%%%%%%%%%%%%%%%%%%%%%%%
%%%%%%%%%%%%%%%%%%%%%%%%%
\section{Conclusion and discussion}
We conjectured the evolution of individual weight matrices should obey Fokker-Planck type equation under training, we tested this in a simple auto-encoder architecture by comparing the output data distributions and obtained good empirical versus theoretical comparison.

The evolution is in fact largely driven by the deterministic piece of the evolution, as in the drift term in \eqref{FP_equation}. The addition of the diffusion term only blurs potential and locally confuses the weight space potential distribution. As discussed earlier, this suggests the evolution of individual weight matrix during training is largely deterministic and follows the Callan-Symanzik equation \eqref{CS_equation}.  

There are a few directions we would like to further pursue in the near future to obtain a more complete understanding of the evolution of the neural network itself during training. 
\begin{itemize}
    \item Coupling between layers (to study other more involved and practically relevant architectures), this would require more theoretical investigations beyond the scope of single layered Fokker-Planck as most architectures are quite convolved. One could wonder how coupling different evolutions across layers can build up distributions across layers and what equation governs the evolution of those.  
    \item Scale up the model to test the equation further and focus on whether curse of dimensionality appears even for a single layer evolution.
    \item Further test how different initializations affect the theoretical predictions.
    \item Use the solution \eqref{Terminal_solution_highd} to predict features of the model before training by solving the equation. Importantly use the Callan-Symanzik equation \eqref{CS_equation} to find the terminal time $T$ at which the probability density is completely controlled by the $\beta$-function, or roughly how we choose to update the weight matrix entries (drift coefficient function). 
\end{itemize}

\section{Acknowledgments}
We would like to thank Jim Halverson, Jin Lee, Nishil Patel, Andrew Saxe, Yedi Zhang for discussion and providing useful references. WB is supported by the Simons Collaboration on celestial holography.

%%%%%%%%%%%%%%%%%%%%%%%%%

\newpage
\bibliography{example_paper}
\bibliographystyle{icml2025}

\end{document}